\documentclass[10pt,twocolumn,letterpaper]{article}

\usepackage[pagenumbers]{cvpr} 

\definecolor{cvprblue}{rgb}{0.21,0.49,0.74}
\usepackage[pagebackref,breaklinks,colorlinks,allcolors=cvprblue]{hyperref}
\usepackage{bm}
\usepackage{pifont}
\usepackage{algorithm}
\usepackage{algorithmic}
\usepackage{placeins}
\usepackage{multirow}
\usepackage{colortbl}

\def\paperID{5149} 
\def\confName{CVPR}
\def\confYear{2025}

\title{DKDM: Data-Free Knowledge Distillation for Diffusion Models with Any Architecture}

\author{
Qianlong Xiang$^{1}$, Miao Zhang$^{1,}$\footnotemark[2], Yuzhang Shang$^{2}$, Jianlong Wu$^{1}$, Yan Yan$^{2}$, Liqiang Nie$^{1,}$\footnotemark[2]\\
$^1$Harbin Institute of Technology, Shenzhen{}
$^2$Illinois Institute of Technology \\
\texttt{\normalsize{\url{https://github.com/qianlong0502/DKDM}}}
}

\begin{document}
\maketitle

\renewcommand{\thefootnote}{\fnsymbol{footnote}} 
\footnotetext[2]{Corresponding authors}

\newcommand{\x}{\boldsymbol{x}}
\newcommand{\xx}{x}
\newcommand{\hx}{\boldsymbol{\hat{x}}}
\newcommand{\thetat}{\boldsymbol{\theta}_T}
\newcommand{\thetas}{\boldsymbol{\theta}_S}
\newcommand{\tthetas}{{\theta}_S}
\newcommand{\hB}{{\hat{\mathcal{B}}}}

\begin{abstract}
Diffusion models (DMs) have demonstrated exceptional generative capabilities across various domains, including image, video, and so on.
A key factor contributing to their effectiveness is the high quantity and quality of data used during training.
However, mainstream DMs now consume increasingly large amounts of data.
For example, training a Stable Diffusion model requires billions of image-text pairs.
This enormous data requirement poses significant challenges for training large DMs due to high data acquisition costs and storage expenses.
To alleviate this data burden, we propose a novel scenario: using existing DMs as data sources to train new DMs with any architecture.
We refer to this scenario as \textbf{D}ata-Free \textbf{K}nowledge Distillation for \textbf{D}iffusion \textbf{M}odels (\textbf{DKDM}), where the generative ability of DMs is transferred to new ones in a data-free manner.
To tackle this challenge, we make two main contributions.
First, we introduce a DKDM objective that enables the training of new DMs via distillation, without requiring access to the data.
Second, we develop a dynamic iterative distillation method that efficiently extracts time-domain knowledge from existing DMs, enabling direct retrieval of training data without the need for a prolonged generative process.
To the best of our knowledge, we are the first to explore this scenario.
Experimental results demonstrate that our data-free approach not only achieves competitive generative performance but also, in some instances, outperforms models trained with the entire dataset.
\end{abstract}

\section{Introduction}
\label{sec:intro}

The advent of Diffusion Models (DMs) \citep{sohl2015deep,ho2020denoising,song2021scorebased} heralds a new era in the generative domain, garnering widespread acclaim for their exceptional capability in producing samples of remarkable quality \citep{dhariwal2021beat,nichol2021improved,rombach2022high}.
These models have rapidly ascended to a pivotal role across a spectrum of generative applications, notably in the fields of image, video and audio \citep{yang2023diffusion,blattmann2023align,huang2023noise2music}.
One reason for their superior performance is their training on large-scale, high-quality datasets. However, this advantage also entails a drawback: training DMs requires substantial storage capacity, as shown in \cref{tab:diff-data}.
For instance, training a Stable Diffusion model necessitates the use of billions of image-text pairs \citep{rombach2022high}.

\begin{figure}[t]
  \centering
  \includegraphics[width=\linewidth]{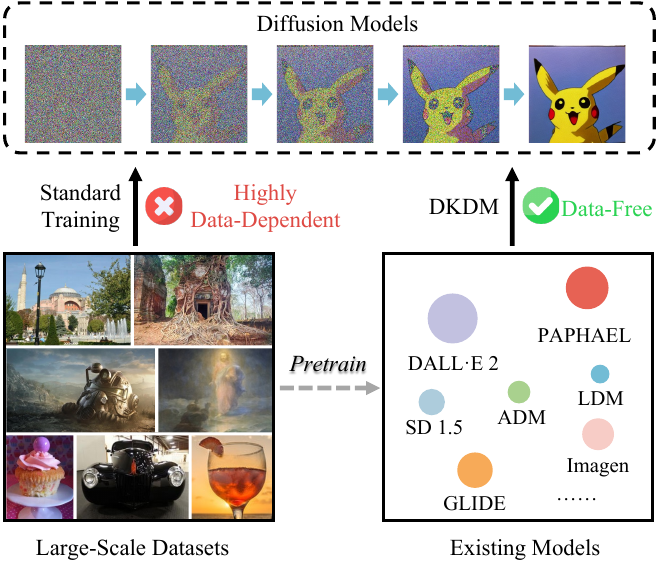}
  \caption{Illustration of our DKDM concept: utilizing pretrained diffusion models to train new ones, thus avoiding the high costs associated with increasingly large datasets.}
  \label{fig:short}
\end{figure}

To alleviate this data burden, considering that numerous pretrained DMs have been trained and released by various organizations, we pose a novel question:
\begin{center}
\textit{
Can we train new diffusion models by using existing pretrained diffusion models as the data source, thereby eliminating the need to access or store any dataset?
}
\end{center}
\cref{fig:short} illustrates the concept of this scenario.
Traditionally, training DMs requires access to large datasets.
In contrast, in this paper, we explore how to utilize existing DMs to train new models without any data.
We formalize this scenario as the \textbf{D}ata-Free \textbf{K}nowledge Distillation for \textbf{D}iffusion \textbf{M}odels (\textbf{DKDM}) paradigm, which aims at transferring the generative ability of the pretrained DMs towards new ones.

\begin{table}[t]
  \centering
    \begin{tabular}{lcc}
    \toprule
    \textbf{Model} & \textbf{\#Param.} & \textbf{\#Images} \\
    \midrule
    GLIDE \cite{nichol2021glide} & 5.0B  & 5.94B \\
    LDM \cite{rombach2022high}  & 1.5B  & 0.27B \\
    DALL·E 2 \cite{ramesh2022hierarchical} & 5.5B  & 5.63B \\
    Imagen \cite{saharia2022photorealistic} & 3.0B  & 15.36B \\
    eDiff-I \cite{balaji2022ediff} & 9.1B  & 11.47B \\
    Stable Diffusion v1.5 \cite{rombach2022high} & 0.9B  & 3.16B \\
    \bottomrule
    \end{tabular}%
  \caption{Comparison of prominent diffusion models on parameter count and training dataset size, sourced from \citet{kang2023scaling}.}
  \label{tab:diff-data}%
\end{table}%

Compared with previous work, our proposed DKDM paradigm imposes strict requirements in \textbf{three aspects}.
\ding{182}~\textbf{Data}.
Previous work usually requires access to datasets to train DMs for purposes such as model compression \citep{yang2023diffusion2,zhang2024laptop} and reducing denoising steps \citep{salimans2022progressive,song2023consistency}. 
In contrast, DKDM mandates that the entire training process must not access any datasets.
This constraint eliminates the need to spend significant time downloading and storing datasets and helps circumvent data privacy issues, especially when training data is not released \citep{zhao2023mobilediffusion}.
\ding{183}~\textbf{Architecture}.
We observe that previous work on knowledge distillation for DMs often initializes student models with the architectures and weights of teacher models, limiting architectural flexibility. One reason for this is to improve performance.
For example, \citet{xie2024distillation} proposed a distillation method for DMs and found that the performance will degrade when the student model is randomly initialized.
On the contrary, DKDM calls for training DMs with any architecture.
\ding{184}~\textbf{Knowledge Form}.
Leveraging deep generative models to synthesize high-quality samples for performance enhancement on downstream tasks is a common practice \cite{azizi2023synthetic,tian2024stablerep,nguyen2024dataset,wu2023datasetdm,li2024genview,rao2024commonit}.
However, we argue that in DKDM, the knowledge form should not be realistic samples, because generating and storing such samples requires enormous space and time.
For instance, to train a model like Stable Diffusion in this way, we would need to use the teacher model to synthesize billions of image-text pairs in advance and then use this massive synthetic dataset to train the new model, which is impractical.
Therefore, the knowledge form in DKDM should be carefully designed.

Based on the above considerations, we summarize the requirements brought by the DKDM paradigm into two key challenges and solve them separately.
The first challenge involves training DMs with any architecture, while not accessing the dataset.
The second challenge involves efficiently designing the knowledge form for distillation, preventing it from becoming the main bottleneck in slowing the training process, as the generation of DMs is inherently slow.
\textbf{For the former}, the optimization objective used in traditional DMs, as described by \citet{ho2020denoising}, is inappropriate due to the absence of the data.
To address this, we specially design a DKDM objective that aligns closely with the original DM optimization objective, while the architecture of the model is no longer limited as in other distillation methods.
\textbf{For the latter}, we observe that compared to realistic samples, time-domain ones corrupted by certain noise are more relevant to the optimization objective for DMs.
Therefore, we define the knowledge form in DKDM as these noisy samples, enabling direct learning from each denoising step of pretrained DMs, without the need for a time-consuming generative process to obtain realistic samples.
In other words, the student model learns from the generative process of the pretrained DMs rather than from their final generative outputs.
Based on this definition, we propose a dynamic iterative distillation method that generates substantial and diverse knowledge to enhance the training of the student.

To sum up, this paper introduces a novel method for training DMs without the need for datasets, by leveraging existing pretrained DMs as the data source.
Experimental results indicate that models trained with our approach demonstrate competitive generative performance.
Furthermore, in some cases, our data-free method even outperforms models trained with the entire dataset.

\section{Preliminaries on Diffusion Models}
\label{pre}
In diffusion models \citep{ho2020denoising}, a Markov chain is defined to add noises to data, and then diffusion models learn the reverse process to generate data from noises.

\textbf{Forward Process.} Given a sample $\x^0 \sim q\left(\x^0\right)$ from the data distribution, the forward process iteratively adds Gaussian noise for $T$ diffusion steps with the predefined noise schedule $\left(\beta_1, \ldots, \beta_T\right)$:
\begin{align}
    q\left(\x^t | \x^{t-1}\right)&=\mathcal{N}\left(\x^t ; \sqrt{1-\beta_t} \x^{t-1}, \beta_t \boldsymbol{I}\right), \\
    q\left(\x^{1: T} | \x^0\right)&=\prod_{t=1}^T q\left(\x^t | \x^{t-1}\right),
\end{align}
until a completely noise $\x^T \sim \mathcal{N}(\mathbf{0}, \boldsymbol{I})$ is obtained. According to \citet{ho2020denoising},
adding noise $t$ times sequentially to the original sample $\x^0$ to generate a \textit{noisy sample} $\x^t$ can be simplified to a one-step calculation as follows:
\begin{equation}
q\left(\x^t | \x^0\right)=\mathcal{N}\left(\x^t ; \sqrt{\bar{\alpha}_t} \x^0,\left(1-\bar{\alpha}_t\right) \boldsymbol{I}\right),
\end{equation}
\begin{equation} \label{eq:xt}
    \x^t=\sqrt{\bar{\alpha}_t} \x^0+\sqrt{1-\bar{\alpha}_t} \boldsymbol{\epsilon},
\end{equation}
where $\alpha_t:=1-\beta_t$, $\bar{\alpha}_t:=\prod_{s=0}^t \alpha_s$ and $\boldsymbol{\epsilon} \sim \mathcal{N}(\mathbf{0}, \boldsymbol{I})$.

\textbf{Reverse Process.} The posterior $q(\x^{t-1} | \x^{t})$ depends on the data distribution, which is tractable conditioned on $\x^0$:
\begin{equation}
q\left(\x^{t-1} | \x^t, \x^0\right)=\mathcal{N}\left(\x^{t-1} ; \tilde{\mu}\left(\x^t, \x^0\right), \tilde{\beta}_t \boldsymbol{I}\right),
\end{equation}
where $\tilde{\mu}_t\left(\x^t, \x^0\right)$ and $\tilde{\beta}_t$ can be calculated by:
\begin{equation}
\tilde{\beta}_t:=\frac{1-\bar{\alpha}_{t-1}}{1-\bar{\alpha}_t} \beta_t,
\end{equation}
\begin{equation} \label{eq:mu1}
\tilde{\mu}_t\left(\x^t, \x^0\right):=\frac{\sqrt{\bar{\alpha}_{t-1}} \beta_t}{1-\bar{\alpha}_t} \x^0+\frac{\sqrt{\alpha_t}\left(1-\bar{\alpha}_{t-1}\right)}{1-\bar{\alpha}_t} \x^t.
\end{equation}
Since $\x^0$ in the data is not accessible during generation, a neural network parameterized by $\boldsymbol{\theta}$ is used for approximation:
\begin{equation}
    p_{\boldsymbol{\theta}}\left(\x^{t-1} | \x^t\right)=\mathcal{N}\left(\x^{t-1} ; \mu_{\boldsymbol{\theta}}\left(\x^t, t\right), \Sigma_{\boldsymbol{\theta}}\left(\x^t, t\right) \boldsymbol{I}\right).
\end{equation}

\textbf{Optimization.} To optimize this network, the variational bound on negative log likelihood $\mathbb{E}[-\log p_{\boldsymbol{\theta}}]$ is estimated by:
\begin{equation} \label{eq:ltm1}
L_{\mathrm{vlb}}=\mathbb{E}_{\x^0,\boldsymbol{\epsilon},t}\left[D_{KL}(q(\x^{t-1}|\x^t,\x^0)||p_{\boldsymbol{\theta}}(\x^{t-1}|\x^t)\right].
\end{equation}
\citet{ho2020denoising} found that predicting $\boldsymbol{\epsilon}$ is a more efficient way when parameterizing $\mu_{\boldsymbol{\theta}}(\x^t,t)$ in practice, which can be derived by \cref{eq:xt,eq:mu1}:
\begin{equation}
\mu_{\boldsymbol{\theta}}\left(\x^t, t\right)=\frac{1}{\sqrt{\alpha_t}}\left(\x^t-\frac{\beta_t}{\sqrt{1-\bar{\alpha}_t}} \boldsymbol{\epsilon}_{\boldsymbol{\theta}}\left(\x^t, t\right)\right).
\end{equation}

Thus, a reweighted loss function is designed as the objective to optimize $L_{\mathrm{vlb}}$:
\begin{equation}
    L_{\mathrm{simple}}=\mathbb{E}_{\x^0, \boldsymbol{\epsilon}, t}\left[\left\|\boldsymbol{\epsilon}-\boldsymbol{\epsilon}_{\boldsymbol{\theta}}\left(\x^t, t\right)\right\|^2\right]. \label{eq:simple}
\end{equation}

\textbf{Improvement.} In original DDPMs, $L_{\mathrm{simple}}$ offers no signal for learning $\Sigma_{\boldsymbol{\theta}}(\x^t,t)$ and \citet{ho2020denoising} fixed it to $\beta_t$ or $\tilde{\beta}_t$. \citet{nichol2021improved} found it to be sub-optimal and proposed to parameterize $\Sigma_{\boldsymbol{\theta}}\left(\x^t, t\right)$ as a neural network whose output $v$ is interpolated as:
\begin{equation} \label{eq:var}
    \Sigma_{\boldsymbol{\theta}}\left(\x^t, t\right)=\exp \left(v \log \beta_t+(1-v) \log \tilde{\beta}_t\right).
\end{equation}

To optimize $\Sigma_{\boldsymbol{\theta}}\left(\x^t, t\right)$, \citet{nichol2021improved} use $L_{\mathrm{vlb}}$, in which a stop-gradient is applied to the $\mu_{\boldsymbol{\theta}}(\x^t,t)$ because it is optimized by $L_{\mathrm{simple}}$. The final hybrid objective is defined as:
\begin{equation} \label{eq:hybrid}
    L_{\mathrm{hybrid}}=L_{\mathrm{simple}} + \lambda L_{\mathrm{vlb}},
\end{equation}
where $\lambda$ is used for balance between the two objectives. The process of training and sampling are guided by \cref{eq:hybrid}, \cf Algorithm~2~and~3 in Sec.~7.

\section{Data-Free Knowledge Distillation for Diffusion Models}
\label{dfkd}

In this section, we introduce a novel paradigm, termed \textbf{D}ata-Free \textbf{K}nowledge Distillation for \textbf{D}iffusion \textbf{M}odels (\textbf{DKDM}).
\cref{paradigm} details the DKDM paradigm, focusing on two principal challenges: the formulation of the optimization objective and the acquisition of knowledge for distillation.
\cref{dkdm-objective} describes our proposed optimization objective tailored for DKDM.
\cref{dynamic-iteration-distillation} details our proposed method for efficient retrieval of knowledge.

\subsection{DKDM Paradigm}
\label{paradigm}
The DKDM paradigm represents a novel scenario for training DMs.
Unlike traditional methods, DKDM aims to leverage existing DMs as the data source to train new ones with any architecture, which eliminates the need for access to large or proprietary datasets.

\begin{figure}[t]
  \centering
  \includegraphics[width=0.5\textwidth]{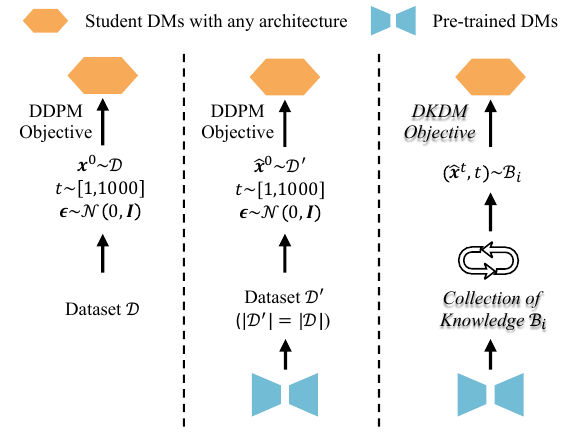}
  
  \begin{subfigure}{.15\textwidth}
    \centering
    \phantom{}
    \caption{Data-Based}
    \label{fig:paradigm-data-based}
  \end{subfigure}
  \hfill
  \begin{subfigure}{.18\textwidth}
    \centering
    \phantom{}
    \caption{Data-Free}
    \label{fig:paradigm-data-free}
  \end{subfigure}
  \hfill
  \begin{subfigure}{.13\textwidth}
    \centering
    \phantom{}
    \caption{Our DKDM}
    \label{fig:paradigm-dkdm}
  \end{subfigure}
  
  \caption{Illustration of our DKDM Paradigm. \textbf{(a)}: standard data-based training of DMs. \textbf{(b)}: a straightforward data-free training approach. \textbf{(c)}: our proposed framework for DKDM.}
  \label{fig:paradigm}
\end{figure}

In standard data-based training of DMs, as depicted in \cref{fig:paradigm-data-based}, a sample $\x^0 \sim \mathcal{D}$ is selected along with a timestep $t \sim \left[1, 1000\right]$ and random noise $\boldsymbol{\epsilon} \sim \mathcal{N}(0, \bm{I})$.
The input $\x^t$ is computed using \cref{eq:xt}, and the denoising network is optimized according to \cref{eq:hybrid} to generate outputs close to $\boldsymbol{\epsilon}$.
However, without dataset access, DKDM cannot obtain training data $(\x^t, t, \boldsymbol{\epsilon})$ to employ this standard method.
A straightforward data-free training approach, depicted in \cref{fig:paradigm-data-free}, involves using DMs pretrained on $\mathcal{D}$ to generate a synthetic dataset $\mathcal{D}^\prime$,
which is then used to train new DMs with varying architectures.
Despite its simplicity, creating $\mathcal{D}^\prime$ is time-intensive and impractical for large datasets.

While data-based training necessitates access to large-scale datasets, data-free training incurs significant costs in generating synthetic datasets.
To address these challenges, we propose an effective and efficient framework for DKDM, outlined in \cref{fig:paradigm-dkdm}, which incorporates a DKDM Objective (described in \cref{dkdm-objective}) and a strategy for collecting knowledge $\mathcal{B}_i$ (detailed in \cref{dynamic-iteration-distillation}).
This framework mitigates the challenges of distillation without datasets and reduces the costs associated with data-free training.

\subsection{DKDM Objective}
\label{dkdm-objective}
Given a dataset $\mathcal{D}$, the original optimization objective for a DM with parameters $\boldsymbol{\theta}$ involves minimizing the KL divergence $\mathbb{E}_{\x^0,\boldsymbol{\epsilon},t}[D_{KL}(q(\x^{t-1}|\x^t,\x^0)\|p_{\boldsymbol{\theta}}(\x^{t-1}|\x^t))]$.
Our proposed DKDM objective comprises two primary goals: (1) eliminating the diffusion posterior $q(\x^{t-1}|\x^t,\x^0)$ and (2) removing the diffusion prior $\x^t \sim q(\x^t | \x^0)$ from the KL divergence, since they both are dependent on $\x^0 \sim \mathcal{D}$.

\textbf{Eliminating the diffusion posterior} $\boldsymbol{q(\xx^{t-1}|\xx^t,\xx^0)}$\textbf{.}
In our framework, we introduce a teacher DM with parameters $\thetat$, trained on dataset $\mathcal{D}$.
This model can generate samples that conform to the learned distribution $\mathcal{D}^\prime$.
Optimized with the objective \cref{eq:hybrid}, the distribution $\mathcal{D}^\prime$ within a well-learned teacher DM closely matches $\mathcal{D}$.
Our goal is for a student, parameterized by $\thetas$, to replicate $\mathcal{D}^\prime$ instead of $\mathcal{D}$, thereby obviating the need for $q$ during optimization.

Specifically, the pretrained teacher DM was optimized via the hybrid objective \cref{eq:hybrid}, which indicates that both the KL divergence $D_{KL}(q(\x^{t-1}|\x^t,\x^0)\|p_{\thetat}(\x^{t-1}|\x^t))$ and the mean squared error $\mathbb{E}_{\x^t, \boldsymbol{\epsilon}, t}[\|\boldsymbol{\epsilon}-\boldsymbol{\epsilon}_{\thetat}(\x^t, t)\|^2]$ are minimized.
Given the similarity in distribution between the teacher model and the dataset, we propose a DKDM objective that optimizes the student model through minimizing $D_{KL}(p_{\thetat}(\x^{t-1}|\x^t)\|p_{\thetas}(\x^{t-1}|\x^t))$ and $\mathbb{E}_{\x^t}[\|\boldsymbol{\epsilon}_{\thetat}(\x^t, t)-\boldsymbol{\epsilon}_{\thetas}(\x^t, t)\|^2]$.
This objective indirectly minimizes $D_{KL}(q(\x^{t-1}|\x^t,\x^0)\|p_{\thetas}(\x^{t-1}|\x^t))$ and $\mathbb{E}_{\x^0, \boldsymbol{\epsilon}, t}[\|\boldsymbol{\epsilon}-\boldsymbol{\epsilon}_{\thetas}(\x^t, t)\|^2]$, despite the inaccessibility of the posterior.
The proposed DKDM objective is as follows:
\begin{equation}
    L_{\mathrm{DKDM}}=L_{\mathrm{simple}}^\prime+\lambda L_{\mathrm{vlb}}^\prime, 
    \label{eq:distill}
\end{equation}
where $L_{\mathrm{simple}}^\prime$ guides the learning of $\mu_{\thetas}$ and $L_{\mathrm{vlb}}^\prime$ optimizes $\Sigma_{\thetas}$, as defined in following equations:
\begin{equation}
    L_{\mathrm{simple}}^\prime = \mathbb{E}_{\x^0,\boldsymbol{\epsilon},t}\left[\|\boldsymbol{\epsilon}_{\thetat}(\x^t,t)-\boldsymbol{\epsilon}_{\thetas}(\x^t,t)\|^2\right],
\end{equation}
\begin{equation}
L_{\mathrm{vlb}}^\prime = \mathbb{E}_{\x^0,\boldsymbol{\epsilon},t}\left[D_{KL}(p_{\thetat}(\x^{t-1}|\x^t)\|p_{\boldsymbol{\tthetas}}(\x^{t-1}|\x^t)\right], 
\end{equation}
where $q(\x^{t-1}|\x^t,\x^0)$ is eliminated whereas the term $\x^t \sim q(\x^t | \x^0)$ remains to be removed.

\textbf{Removing the diffusion prior} $\boldsymbol{q(\xx^t | \xx^0)}$\textbf{.}
Considering the generative ability of the teacher model, we utilize it to generate $\hx^t$ as a substitute for $\x^t \sim q(\x^t | \x^0)$.
We define a reverse diffusion step $\hx^{t-1} \sim p_{\thetat}(\hx^{t-1}|\x^t)$ through the equation $\hx^{t-1}=g_{\thetat}(\x^{t}, t)$.
Next, we represent a sequence of $t$ reverse diffusion steps starting from $T$ as $G_{\thetat}(t)$.
Note that $G_{\thetat}(0)=\boldsymbol{\epsilon}$ where $\boldsymbol{\epsilon} \sim \mathcal{N}(0, \bm{I})$.
For instance, $G_{\thetat}(2)$ yields $\hx^{T-2}=g_{\thetat}(g_{\thetat}(\boldsymbol{\epsilon}, T), T-1)$.
Consequently, $\hx^t$ is obtained by $\hx^t=G_{\thetat}(T-t)$ and the objectives $L_{\mathrm{simple}}^\prime$ and $L_{\mathrm{vlb}}^\prime$ are reformulated as follows:
\begin{equation}
    L_{\mathrm{simple}}^\prime = \mathbb{E}_{\hx^t, t}\left[\|\boldsymbol{\epsilon}_{\thetat}(\hx^t,t)-\boldsymbol{\epsilon}_{\thetas}(\hx^t,t)\|^2\right], \label{eq:simples}
\end{equation}
\begin{equation}
L_{\mathrm{vlb}}^\prime = \mathbb{E}_{\hx^t,t}\left[D_{KL}(p_{\thetat}(\hx^{t-1}|\hx^t)\|p_{\boldsymbol{\tthetas}}(\hx^{t-1}|\hx^t)\right].
\label{eq:vlbs}
\end{equation}

By this formulation, the need for $\x^0$ in $L_{\mathrm{DKDM}}$ is removed by naturally leveraging the generative ability of the teacher.
Optimized by the proposed $L_{\mathrm{DKDM}}$, the student progressively learns the entire reverse diffusion process from the teacher without reliance on the source datasets.

However, the removal of the diffusion posterior and prior in the DKDM objective introduces a significant bottleneck, resulting in notably slow learning rates.
As depicted in \cref{fig:paradigm-data-based}, standard training for DMs enables straightforward acquisition of noisy samples $\x_i^{t_i}$ at an arbitrary diffusion step $t \sim [1,T]$ using \cref{eq:xt}.
These samples are compiled into a batch $\mathcal{B}_j=\{\x_i^{t_i}\}$, with $j$ representing the training iteration.
Conversely, our DKDM objective requires obtaining a noisy sample $\hx_i^t=G_{\thetat}(T-t_i)$ through $T-t_i$ denoising steps.
Consequently, by considering the denoising steps as the primary computational expense, the worst-case time complexity of assembling a batch $\hB_j=\{\hx_i^{t_i}\}$ for distillation is ${\mathcal{O}(Tb)}$, where $b$ denotes the batch size.
This complexity significantly hinders the training process.
To address this issue, we introduce a method called dynamic iterative distillation, detailed in \cref{dynamic-iteration-distillation}.

\subsection{Efficient Collection of Knowledge}
\label{dynamic-iteration-distillation}
\begin{figure*}[t]
\centering
\includegraphics[width=\linewidth]{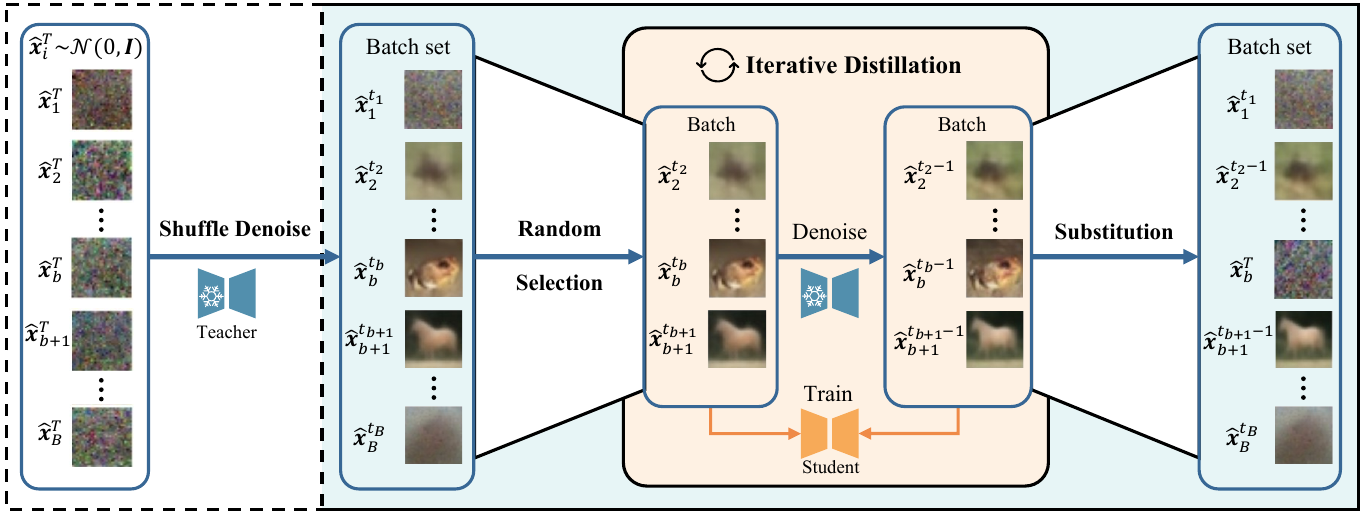}
\caption{
\textbf{Dynamic Iterative Distillation:}
An enlarged batch set is initially constructed by sampling from a Gaussian distribution.
Next, shuffle denoise is applied, wherein each sample is denoised random times.
A batch is then randomly selected from this enlarged set for training the student with the denoised results substituting for their counterparts in the batch set.
This process is repeated iteratively.
}
\label{fig:framework}
\end{figure*}

In this section, we present our efficient strategy for gathering knowledge for distillation, illustrated in \cref{fig:framework}.
We begin by introducing a basic iterative distillation method that allows the student to learn from the teacher at each denoising step, instead of requiring the teacher to denoise multiple times within every training iteration to create a batch of noisy samples.
Subsequently, to enhance the diversity of noise levels within the batch samples, we develop an advanced method termed shuffled iterative distillation, which allows the student to learn denoising patterns across varying time steps.
Lastly, we refine our approach to dynamic iterative distillation, significantly augmenting the diversity of data in the batch. This adaptation ensures that the student acquires knowledge from a broader array of samples over time, avoiding repetitive learning from identical samples.

\textbf{Iterative Distillation.}
We introduce a method called iterative distillation, which closely aligns the optimization process with the generation procedure.
In this approach, the teacher model consistently denoises, while the student model continuously learns from this denoising.
Each output from the teacher's denoising step is incorporated into some batch for optimization, ensuring the student model learns from every output.
Specifically, during each training iteration, the teacher performs $g_{\thetat}(\x^t,t)$, which is a single-step denoising, instead of $G_{\thetat}(t)$, which would involve $t$-step denoising.
Initially, a batch $\hB_1=\{\hx_{i}^{T}\}$ is formed from a set of sampled noises $\hx_{i}^{T} \sim \mathcal{N}(\boldsymbol{0},\boldsymbol{I})$.
After one step of distillation, the batch $\hB_2=\{\hx_{i}^{T-1}\}$ is used for training.
This process is iterated until $\hB_{T}=\{\hx_{i}^{1}\}$ is reached, indicating that the batch has nearly become real samples with no noise.
The cycle then restarts with the resampling of noise to form a new batch $\hB_{T+1}=\{\hx_{i}^{T}\}$.
This method allows the teacher model to provide an endless stream of data for distillation.
To further improve the diversity of the synthetic batch $\hB_j=\{\hx_{i}^{t_i}\}$, we investigate it from the perspectives of noise level $t_i$ and sample $\hx_i$.

\textbf{Shuffled Iterative Distillation.}
Unlike the standard data-based training, the $t$ values in an iterative distillation batch remain the same and do not follow a uniform distribution, resulting in significant instability during distillation.
To mitigate this issue, we integrate a method termed shuffle denoise into our iterative distillation.
Initially, a batch $\hB_0^s=\{\hx_i^{T}\}$ is sampled from a Gaussian distribution.
Subsequently, each sample undergoes random denoising steps, resulting in $\hB_1^s=\{\hx_i^{t_i}\}$, with $t_i$ following a uniform distribution.
This batch, $\hB_1^s$, then initiates the iterative distillation process.
By enhancing the diversity in the $t_i$ values within the batch, this method balances the impact of different $t$ values during distillation.

\textbf{Dynamic Iterative Distillation.}
There is a notable distinction between standard training and iterative distillation regarding the flexibility in batch composition.
Consider two samples, $\hx_1$ and $\hx_2$, within a batch without differentiating their noise level.
During standard training, the pairing of $\hx_1$ and $\hx_2$ is entirely random.
Conversely, in iterative distillation, batches containing $\hx_1$ almost always include $\hx_2$.
This departure from the principle of independent and identically distributed samples in a batch can potentially diminish the model's generalization ability.

\begin{algorithm}[h]
    \caption{Dynamic Iterative Distillation}
    \label{alg:practicaldfkd}
    \begin{algorithmic}[1]
    \REQUIRE $\hB_0^+=\{\hx_i^{T}\}$
    \STATE Get $\hB_1^+=\{\hx_i^{t_i}\}$ with shuffle denoise, $j=0$
    \REPEAT
        \STATE $j=j+1$
        \STATE get $\hB_j^s$ from $\hB_j^+$ through random selection
        \STATE compute $L_{\mathrm{simple}}^\star$ using \cref{eq:didsimples}
        \STATE compute $L_{\mathrm{vlb}}^\star$ using \cref{eq:var,eq:didvlbs}
        \STATE $\text{take a gradient descent step on} \: 
        \nabla_{\boldsymbol{\theta}}L_{\mathrm{DKDM}}$
        \STATE update $\hB_{j+1}^+$
        \UNTIL {converged}
    \end{algorithmic}
\end{algorithm}

To better align the distribution of the denoising data with that of the standard training batch, we propose a method named dynamic iterative distillation.
As shown in \cref{fig:framework}, this method employs shuffle denoise to construct an enlarged batch set $\hB_1^+=\{\hx_i^{t_i}\}$, where size $|\hB_{j}^{+}|=\rho T|\hB_{j}^{s}|$, where $\rho$ is a scaling factor.
During distillation, a subset $\hB_j^s$ is sampled from $\hB_j^+$ through random selection for optimization.
The one-step denoised samples replace their counterparts in $\hB_{j+1}^+$.
This method only has a time complexity of ${\mathcal{O}(b)}$ and significantly improves distillation performance.
The final DKDM objective is defined as:
\begin{equation}
    L_{\mathrm{DKDM}}^\star=L_{\mathrm{simple}}^\star+\lambda L_{\mathrm{vlb}}^\star, \label{eq:diddistill}
\end{equation}
\begin{equation}
    L_{\mathrm{simple}}^\star = \mathbb{E}_{(\hx^t,t) \sim \hB^+}\left[\|\boldsymbol{\epsilon}_{\thetat}(\hx^t,t)-\boldsymbol{\epsilon}_{\thetas}(\hx^t,t)\|^2\right], \label{eq:didsimples}
\end{equation}
\begin{equation}
L_{\mathrm{vlb}}^\star = \mathbb{E}_{(\hx^t,t) \sim \hB^+}\left[D_{KL}(p_{\thetat}(\hx^{t-1}|\hx^t)\|p_{\boldsymbol{\tthetas}}(\hx^{t-1}|\hx^t)\right], \label{eq:didvlbs}
\end{equation}
where $\hx^t$ and $t$ are produced by our proposed dynamic iterative distillation. The complete algorithm is detailed in \cref{alg:practicaldfkd}.

\begin{table*}[t]
  \centering
    \begin{tabular}{cccccccccc}
    \toprule
      \multirow{2}[4]{*}{\textbf{Method}}  & \multicolumn{3}{c}{\textbf{CIFAR10 32x32}} & \multicolumn{3}{c}{\textbf{CelebA 64x64}} & \multicolumn{3}{c}{\textbf{ImageNet 32x32}} \\
\cmidrule{2-10}          & \textbf{IS↑} & \textbf{FID↓} & \textbf{sFID↓} & \textbf{IS↑} & \textbf{FID↓} & \textbf{sFID↓} & \textbf{IS↑} & \textbf{FID↓} & \textbf{sFID↓} \\
    \midrule
    Teacher & 9.52  & 4.45  & 7.09  & 3.08  & 4.43  & 6.10  & 13.63  & 4.67  & 4.03  \\
    Data-Based Training & 8.73  & 7.84  & 7.38  & 3.04  & 5.39  & 7.23  & 9.99  & 10.56  & 5.24  \\
    \midrule
    Data-Limited Training (20\%) & \underline{8.49}  & \underline{9.76}  & \textbf{11.30 } & 2.86  & 9.52  & 11.55  & \underline{10.48}  & \underline{12.62}  & 9.70  \\
    Data-Limited Training (15\%) & 8.44  & 11.07  & 12.47  & 2.84  & 9.60  & 11.48  & 10.43  & 13.50  & 10.76  \\
    Data-Limited Training (10\%) & 8.40  & 11.06  & 11.98  & \textbf{3.04 } & \underline{8.20}  & \underline{10.65}  & 10.39  & 14.23  & 12.63  \\
    Data-Limited Training (5\%) & 8.39  & 10.91  & 11.99  & 2.86  & 9.64  & 11.27  & \underline{10.48}  & 13.63  & 10.62  \\
    Data-Free Training (0\%) & 8.28  & 12.06  & 13.23  & 2.87  & 10.66  & 12.71  & 10.47  & 13.20  & \underline{9.56}  \\
    \midrule
    Dynamic Iterative Distillation (Ours) & \textbf{8.60} & \textbf{9.56} & \underline{11.77}  & \underline{2.91}  & \textbf{7.07} & \textbf{8.78} & \textbf{10.50} & \textbf{11.33} & \textbf{4.80} \\
    \bottomrule
    \end{tabular}%
  \caption{Pixel-space performance comparison between data-limited training, data-free training and our dynamic iterative distillation on CIFAR10 $32 \times 32$ \citep{cifar10}, CelebA $64 \times 64$ \citep{celeba} and ImageNet $32 \times 32$ \cite{imagenet32}. The term (P\%) denotes the percentage of real data included in the synthetic dataset. The best performance is indicated by \textbf{boldface}, while the second-best is denoted by \underline{underlining}. Results from the ‘Teacher’ and ‘Data-Based Training’ are provided for reference only and are not included in the comparison.}
  \label{tab:pixel-result}%
\end{table*}%

\section{Experiments}
\label{experiment}
This section presents a series of experiments designed to validate the efficacy of our proposed dynamic iterative distillation.
In \cref{experiment-setting}, we introduce our experimental setting and establish relevant baselines for comparative analysis from a data-centric perspective.
\cref{main-result} provides a comparison between these baselines and our method, assessing performance separately in pixel and latent spaces.
We also demonstrate the capability of our approach to train models across different architectures.
Lastly, Section \ref{ablation} includes an ablation study to solidify the validation of our method.

\subsection{Experiment Setting}
\label{experiment-setting}
\textbf{Datasets, teachers and students.}
The training of high-resolution diffusion models typically requires substantial time, so these models are often developed in latent space to expedite the process \cite{rombach2022high}.
To assess our method, we conduct experiments in both pixel and latent spaces, focusing on low and high-resolution generation, respectively.
\begin{itemize}
    \item \textbf{Pixel space.} We utilize three pretrained DMs as teacher models, following the configurations introduced by \citet{ning2023input}. These models were trained separately on CIFAR10 at a resolution of $32 \times 32$ \citep{cifar10}, CelebA at $64 \times 64$ \citep{celeba} and ImageNet at $32 \times 32$ \cite{imagenet32}.
    \item \textbf{Latent space.} We adopt two different DMs as teacher models, adhering to the configurations proposed by \citet{rombach2022high}. These models were trained on CelebA-HQ $256 \times 256$ \cite{karras2018progressive}
    and FFHQ $256 \times 256$ \cite{karras2019style}.
    It is important to note that the pre-trained models in the latent space were typically trained using a simpler loss function, denoted as $L_{\mathrm{simple}}$~\eqref{eq:simple}, without incorporating the KL divergence $L_{\mathrm{vlb}}$~\eqref{eq:ltm1}. For our experiments, we adopt $L_{\mathrm{simple}}^\star$~\eqref{eq:didsimples} as the DKDM objective. This approach allows us to investigate the effectiveness of dynamic iterative distillation under different training conditions.
\end{itemize}
All the teacher models employ Convolutional Neural Networks (CNNs). For the student models, we maintain the same architecture but reduce the scale. Additionally, we conduct cross-architecture experiments between CNN-based and ViT-based (Vision Transformer \cite{dosovitskiy2021an,peebles2023scalable}) DMs on CIFAR10. Details of the architecture are listed in Sec.~8.

\textbf{Metrics.}
The distance between the generated samples
and the reference samples can be estimated by the \textbf{F}réchet \textbf{I}nception \textbf{D}istance (FID) score \citep{heusel2017gans}.
In our experiments, we utilize the FID score as the primary metric for evaluation.
Additionally, we report sFID \citep{charlie2021generating} and Inception Score (IS) \cite{salimans2016improved} as secondary metrics.
Following previous work \citep{ho2020denoising,nichol2021improved,ning2023input}, we generate 50K samples for DMs, and we use the full training set in the corresponding dataset to compute the metrics.
Without additional contextual states, all the samples are generated through 50 Improved DDPM sampling steps \citep{nichol2021improved} in pixel space and 200 DDIM sampling steps \citep{song2021denoising} in latent space.
All of our metrics are calculated by ADM TensorFlow evaluation suite \citep{dhariwal2021beat}.

\textbf{Baselines.}
As DKDM is a new paradigm proposed in this paper, traditional distillation methods are not suitable to serve as baselines.
Therefore, we establish two kinds of baselines from a data-centric perspective.
\begin{itemize}
    \item \textbf{Data-Free Training}, which is depicted in \cref{fig:paradigm-data-free}, involves a teacher model generating a large quantity of high-quality synthetic samples, matching the size of the original dataset. These synthetic samples form the training set $\mathcal{D}^\prime$ for the student models, which are initialized randomly and trained according to the standard procedure, \cf Algorithm~2 in Sec.~7. Details about our synthetic datasets can be found in \cref{tab:dataset-image}.
    \item \textbf{Data-Limited Training} integrates a fixed proportion (ranging from 5\% to 20\%) of the original dataset samples with the synthetic dataset $\mathcal{D}^\prime$ used in data-free training. It facilitates a comparative analysis between our purely data-free method and those able to partially access to the original dataset.
\end{itemize}
Additionally, we also report performance of data-based training, illustrated in Figure \ref{fig:paradigm-data-based}, which serves as an upper performance limit for our analysis.

\begin{table}[t]
  \centering
  \scalebox{0.95}{
    \begin{tabular}{ccccc}
    \toprule
    \multirow{2}[3]{*}{\textbf{Method}} & \multicolumn{2}{c}{\textbf{CelebA-HQ 256}} & \multicolumn{2}{c}{\textbf{FFHQ 256}} \\
\cmidrule{2-5}          & \textbf{FID↓} & \textbf{sFID↓} & \textbf{FID↓} & \textbf{sFID↓} \\
    \midrule
    Teacher & 5.69  & 10.02  & 5.93  & 7.52  \\
    Data-Based & 9.09  & 12.10  & 8.91  & 8.75  \\
    \midrule
    Data-Limited (20\%) & \underline{14.49}  & 17.08  & \underline{15.43}  & \underline{12.25}  \\
    Data-Limited (15\%) & 14.89  & \underline{16.98}  & 16.02  & 12.48  \\
    Data-Limited (10\%) & 15.23  & 17.53  & 16.00  & 12.47  \\
    Data-Limited (5\%) & 15.07  & 17.64  & 15.86  & 12.56  \\
    Data-Free (0\%) & 15.36  & 17.56  & 16.32  & 12.75  \\
    \midrule
    Ours  & \textbf{8.69} & \textbf{12.50} & \textbf{11.53} & \textbf{10.29} \\
    \bottomrule
    \end{tabular}%
  }
  \caption{Latent-space performance comparison between data-limited training, data-free training and our dynamic iterative distillation on CelebA-HQ $256 \times 256$ \cite{karras2018progressive} and FFHQ $256 \times 256$ \cite{karras2019style}. The term (P\%) denotes the percentage of real data included in the synthetic dataset. The best performance is indicated by \textbf{boldface}, while the second-best is denoted by \underline{underlining}. Results from the ‘Teacher’ and ‘Data-Based Training’ are provided for reference only and are not included in the comparison.}
  \label{tab:latent-result}%
\end{table}%

\begin{table}[t]
  \centering
    \scalebox{0.96}{
    \begin{tabular}{ccc}
    \toprule
    Dataset & \#Images & Method \\
    \midrule
    \rowcolor[rgb]{ .949,  .949,  .949} \multicolumn{3}{c}{\textit{Pixel Space}} \\
    \midrule
    CIFAR10 32 \citep{cifar10} & 50,000 & IDDPM-1000 \cite{nichol2021improved} \\
    CelebA 64 \citep{celeba} & 202,599 & IDDPM-100 \cite{nichol2021improved} \\
    ImageNet 32 \citep{imagenet32} & 1,281,167 & IDDPM-100 \cite{nichol2021improved} \\
    \midrule
    \rowcolor[rgb]{ .949,  .949,  .949} \multicolumn{3}{c}{\textit{Latent Space}} \\
    \midrule
    CelebA-HQ 256 \cite{karras2018progressive} & 25,000 & DDIM-100 \cite{song2021denoising} \\
    FFHQ 256 \cite{karras2019style} & 60,000 & DDIM-100 \cite{song2021denoising} \\
    \bottomrule
    \end{tabular}%
    }
  \caption{Image counts and generation methods for our baselines, which match the training set sizes of their respective teachers. The notation $*-N$ indicates that the synthetic dataset is generated using $N$ sampling steps with the $*$ method.}
  \label{tab:dataset-image}%
\end{table}%

\subsection{Main Results}
\label{main-result}
\begin{figure*}[t]
\begin{minipage}{.57\textwidth}

\newcommand{\tmpwidth}{127pt}
\begin{figure}[H]
\centering
\begin{subfigure}{.53\linewidth}
    \includegraphics[height=\tmpwidth]{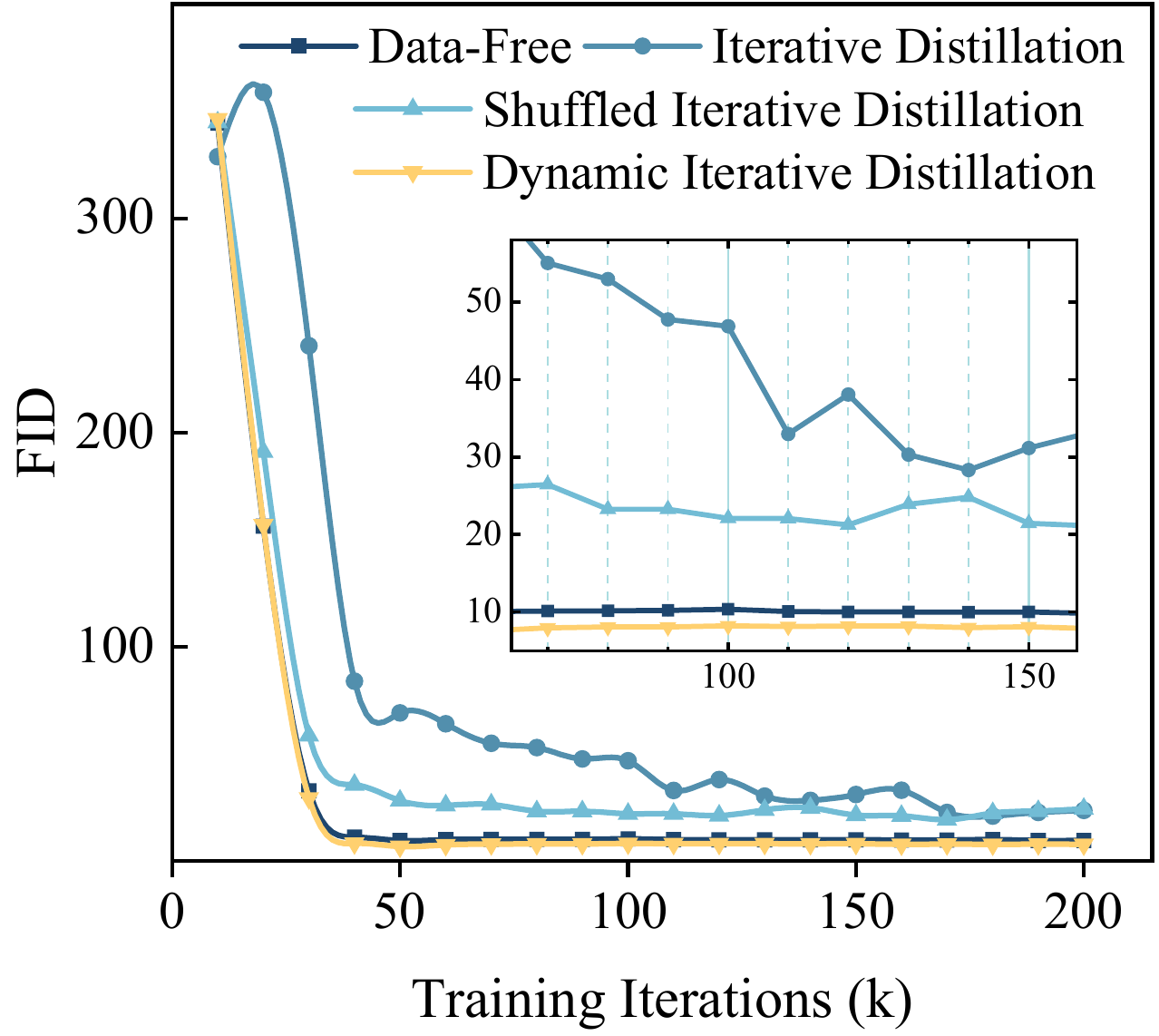}
    \caption{Ablation}
    \label{fig:strategy}
\end{subfigure}
\begin{subfigure}{.46\linewidth}
    \includegraphics[height=\tmpwidth]{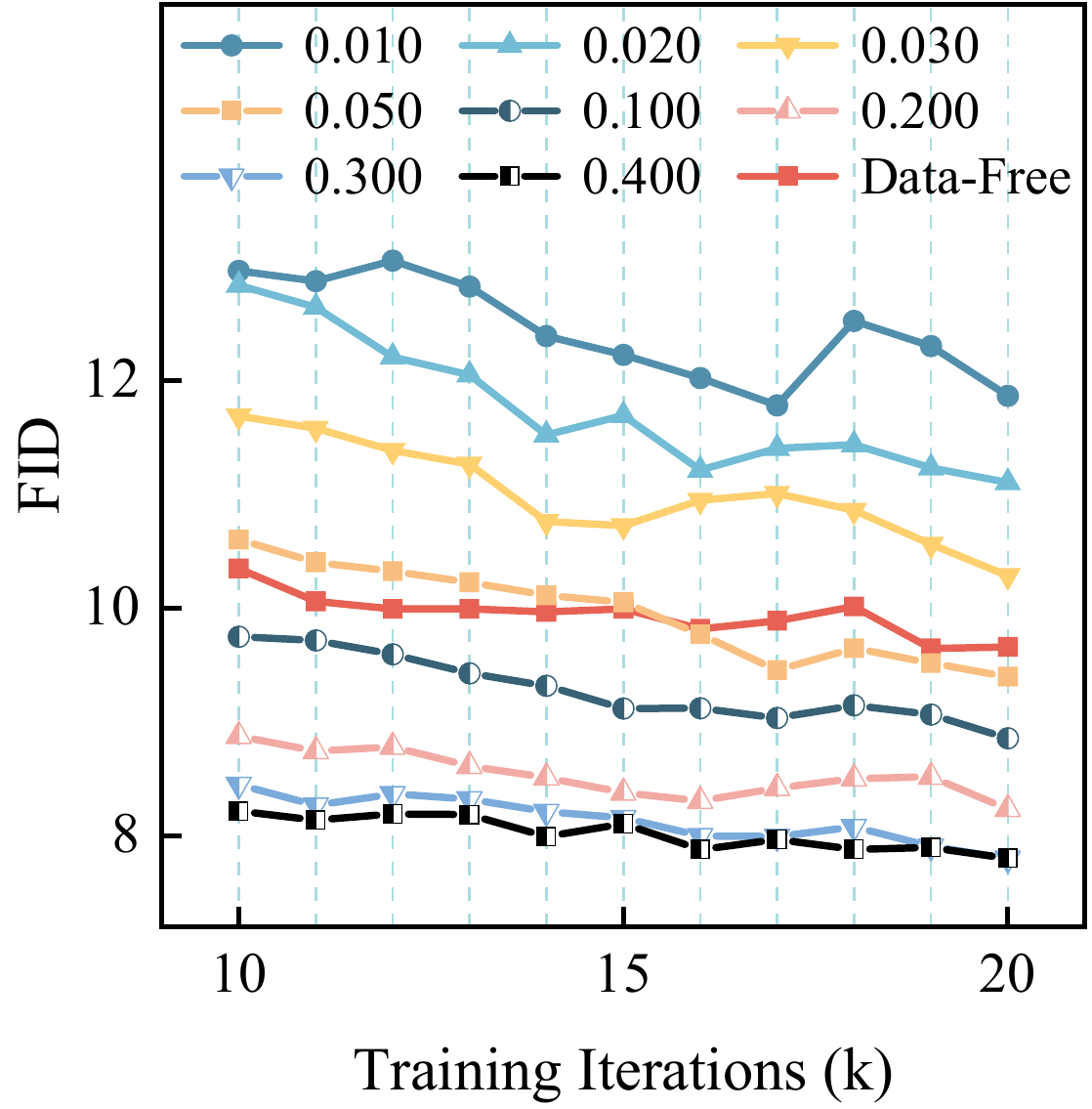}
    \caption{Effect of $\rho$}
    \label{fig:rho}
\end{subfigure}
\caption{FID scores of analytical experiments on CIFAR10. \textbf{(a)}: Ablation on dynamic iterative distillation with $\rho=0.4$. \textbf{(b)}: Effect of different $\rho$.}
\label{fig:ablation}
\end{figure}

\end{minipage}
\hfill
\begin{minipage}{.405\textwidth}

\begin{figure}[H]
\centering
\includegraphics[width=.85\textwidth]{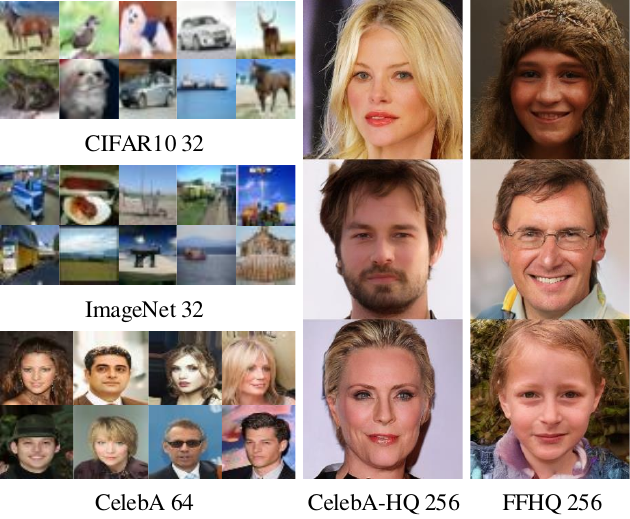}
\caption{Selected samples generated by our student models across five datasets.}
\label{fig:samples}
\end{figure}

\end{minipage}

\end{figure*}

\textbf{Effectiveness.}
\cref{tab:pixel-result,tab:latent-result} present the performance comparison between our dynamic iterative distillation method and baseline models in pixel and latent spaces, respectively.
Our trained students consistently outperform baselines across various datasets and metrics, demonstrating the efficacy of our proposed DKDM objective and dynamic iterative distillation approach.
These results validate our initial hypothesis posited in \cref{sec:intro} and confirm that leveraging existing diffusion models to train new ones is an effective strategy to mitigate the costs associated with large-scale datasets.
Additionally, we observe instances where our method outperforms traditional data-based training approaches, exemplified by the IS score on CIFAR10 and the FID score on CelebA-HQ.
This outcome indicates that, to some extent, neural networks face challenges in learning the complex reverse diffusion processes inherent in data-based training, whereas the knowledge from pretrained teacher models is easier to learn.
This insight further highlights an additional benefit: our method not only reduces the reliance on extensive datasets but also potentially yields models with superior performance.
Moreover, we found our method only consumes minor extra GPU memory while achieving faster training speed in latent space, \cf Sec.~10.
Some generated results are visualized in \cref{fig:samples}.

\begin{table}[t]
    \begin{center}
    \begin{tabular}{p{4.2cm}p{1.2cm}<{\centering}p{1.2cm}<{\centering}}
    \toprule
    & CNN T. & ViT T. \\
    \midrule
    \multicolumn{1}{l}{\textbf{CNN S.}}& & \\
    Data-Free Training & 9.64 & 44.62 \\
    Dynamic Iterative Distillation & \textbf{6.85} & \textbf{13.17} \\
    \midrule
    \multicolumn{1}{l}{\textbf{ViT S.}} & &\\
    Data-Free Training & \textbf{17.11} & 63.15 \\
    Dynamic Iterative Distillation & \textbf{17.11} & \textbf{17.86} \\
    \bottomrule
    \end{tabular}
    \end{center}
    \caption{FID scores on CIFAR10 for cross-architecture distillation between CNN and ViT models. The FID score for the CNN teacher model is 4.45, and that of the ViT teacher is 11.30. Abbreviations used: ‘T.’ stands for teacher, ‘S.’ stands for student.}
    \label{tab:cross}
\end{table}

\textbf{Cross-Architecture Distillation.} 
Our method transcends specific model architectures, enabling distillation from CNN-based DMs to ViT-based ones and vice versa.
As shown in \cref{tab:cross}, our method effectively facilitates cross-architecture distillation, yielding superior performance compared to baselines.
Additionally, our results suggest that CNNs are more effective as compressed DMs.

\subsection{Ablation Study}
\label{ablation}
To validate our approach, we tested the FID score of our progressively designed methods, including iterative, shuffled iterative, and dynamic iterative distillation, over 200K training iterations without early stop.
The results, shown in \cref{fig:strategy}, demonstrate that our dynamic iterative distillation strategy not only converges more rapidly but also delivers superior performance.
The convergence curve for our method closely matches that of the baseline, which confirms the effectiveness of the DKDM objective in alignment with the standard optimization objective \cref{eq:hybrid}.

Further experiments explored the effects of varying $\rho$ on the performance of dynamic iterative distillation.
As depicted in \cref{fig:rho}, higher $\rho$ values enhance the distillation process up to a point, beyond which performance gains diminish.
This outcome supports our hypothesis that dynamic iterative distillation enhances batch construction flexibility, thereby improving distillation efficiency.
Beyond a certain level of flexibility, further increases in $\rho$ yield no significant benefit to the distillation process.
For information regarding GPU memory consumption with varying $\rho$, \cf Sec.~10.
Additional discussion and analytical experiments are available in Secs.~11~and~12.

\section{Related Work}
\textbf{Knowledge Distillation for Diffusion Models.}
Knowledge Distillation (KD) \cite{hinton2015distilling,gou2021knowledge,li2023mask,yang2022cross} is an effective method for transferring the capabilities from teacher models to students for model compression \cite{li2020local,sanh2019distilbert,jiao2019tinybert,rao2024parameter,rao2023dynamic,park2019relational,yang2024clip}.
In the context of diffusion models, KD is usually adopted to accelerate the inherently slow generation process, which involves multiple sampling steps.
Approaches in this domain generally fall into two categories: 1) reducing model size \cite{yang2023diffusion2,zhang2024laptop}
and 2) decreasing sampling steps
\cite{luhman2021knowledge,salimans2022progressive,song2023consistency,gu2023boot,sauer2023adversarial,sauer2024fast,meng2023distillation,xie2024distillation}.
The first strategy focuses on distilling smaller models to reduce inference time, while the second distills the multi-step sampling behavior of teacher models into fewer steps for the student, thereby accelerating generation.
Distinct from these conventional acceleration-oriented KD methods, our approach shifts focus towards the data perspective, aiming to mitigate the extensive data requirements of training diffusion models by distilling knowledge from teacher models to randomly initialized student models in a data-free manner.
Among existing methods, the BOOT method proposed by \citet{gu2023boot} employs a data-free knowledge distillation approach to reduce sampling steps and is most closely related to our work.
However, their primary difference lies in the architecture of the student model.
The BOOT method retains both the structure and weights from the teacher, thereby limiting the flexibility of the student.
In contrast, our method permits any student architecture.

\textbf{Data-Free Knowledge Distillation.}
Traditional data-free knowledge distillation typically transfers knowledge from a slow teacher model to a lightweight student without needing access to the original training dataset, addressing privacy concerns.
Early methods optimized randomly initialized noise to produce synthetic data \cite{yin2020dreaming,nayak2019zero,binici2022robust} for distillation.
Owing to the slow nature of this optimization, subsequent studies have employed generative models to synthesize training data \cite{chen2019data,choi2020data,fang2022up,fang2021contrastive,luo2020large,micaelli2019zero,yoo2019knowledge,yu2023data}.
These efforts primarily distilled knowledge for non-generative models, such as classification networks.
In contrast, this paper focuses on the distillation of generative diffusion models themselves.
We deeply dive into the generation mechanism of diffusion models and design an effective and efficient method to produce synthetic data for distillation.

\section{Conclusion}
In this paper, we aim at addressing rapidly increasing cost associated with the demand for large-scale datasets in training diffusion models.
To mitigate this data burden, we introduce \textbf{D}ata-Free \textbf{K}nowledge Distillation for \textbf{D}iffusion \textbf{M}odels (\textbf{DKDM}), a novel scenario that utilizes pretrained diffusion models to train new ones with any architecture, while not requiring access to the original training dataset.
To achieve this, we carefully design a DKDM objective and dynamic iterative distillation method, which separately guarantees effectiveness and efficiency in the training process of the student model.
To the best of our knowledge, we are the first to explore this scenario and make initial efforts.
Our experiments show superior performance across five datasets, including both pixel and latent spaces.
Furthermore, in some cases, our data-free method even outperforms models trained with the entire dataset.
This offers a more efficient direction for training diffusion models from a data perspective, providing a valuable insight for future advancements.

\FloatBarrier

{
    \small
    \bibliographystyle{ieeenat_fullname}
    \bibliography{main}
}

\end{document}